\documentclass[lettersize,journal]{IEEEtran}
\usepackage[dvipsnames, svgnames, x11names]{xcolor}
\usepackage{dblfloatfix} 

\usepackage{epstopdf}
\usepackage{graphicx}
\usepackage{hyperref}
\usepackage[utf8]{inputenc}
\usepackage{longtable}
\usepackage{makecell}
\usepackage{mathrsfs}
\usepackage{multicol}
\usepackage{multirow}
\usepackage{textcomp}
\usepackage{url} 
\usepackage{amssymb}
\usepackage{pgfplots}
\usepackage{amsthm}
\usepackage{amsmath,amssymb,amsfonts}
\usepackage[ruled,vlined]{algorithm2e}
\ifCLASSOPTIONcompsoc
  \usepackage[nocompress]{cite}
\else
  \usepackage{cite}
\fi

\newtheorem{theorem}{Theorem}[section]
\newtheorem{corollary}[theorem]{Corollary}
\newtheorem{lemma}[theorem]{Lemma}

\newtheorem{a.theorem}[theorem]{Theorem}
\newtheorem{a.corollary}[theorem]{Corollary}
\newtheorem{a.lemma}[theorem]{Lemma}

\begin{document}

\title{Contrastive Learning for Semi-Supervised Deep Regression with Generalized Ordinal Rankings from Spectral Seriation}

\author{Ce Wang, Weihang Dai, Hanru Bai,
        Xiaomeng Li,~\IEEEmembership{Senior Member,~IEEE}
\IEEEcompsocitemizethanks{
\IEEEcompsocthanksitem Ce Wang, Weihang Dai, Xiaomeng Li are with the Department of Electronic and Computer Engineering, The Hong Kong University of Science and Technology, Hong Kong, Ce Wang is also affiliated with the School of Science, Sun Yat-sen University, Shenzhen
510275, China (e-mail: wangc79@mail.sysu.edu.cn; eexmli@ust.hk).
\IEEEcompsocthanksitem Hanru Bai is with the Institute of Science and Technology for Brain-Inspired Intelligence, Fudan University, Shanghai (e-mail: hrbai23@m.fudan.edu.cn).
\IEEEcompsocthanksitem Ce Wang and Weihang Dai contributed equally to this work.
\IEEEcompsocthanksitem Corresponding author: Xiaomeng Li.
\IEEEcompsocthanksitem This work was supported in part by the National Natural Science Foundation of China (NSFC) under Grant No. 62301532 and Grant No. 62306254, in part by the Hong Kong Innovation and Technology Fund under Grant No. GHP/124/22, and in part by the Natural Science Foundation of Jiangsu Province under Grant No. BK20230282.
}
}

\markboth{Journal of \LaTeX\ Class Files,~Vol.~14, No.~8, August~2021}%
{Shell \MakeLowercase{\textit{et al.}}: A Sample Article Using IEEEtran.cls for IEEE Journals}


\maketitle

\begin{abstract}
Contrastive learning methods enforce label distance relationships in feature space to improve representation capability for regression models. However, these methods highly depend on label information to correctly recover ordinal relationships of features, limiting their applications to semi-supervised regression. In this work, we extend contrastive regression methods to allow unlabeled data to be used in the semi-supervised setting, thereby reducing the dependence on costly annotations. Particularly we construct the feature similarity matrix with both labeled and unlabeled samples in a mini-batch to reflect inter-sample relationships, and an accurate ordinal ranking of involved unlabeled samples can be recovered through spectral seriation algorithms if the level of error is within certain bounds. The introduction of labeled samples above provides regularization of the ordinal ranking with guidance from the ground-truth label information, making the ranking more reliable. To reduce feature perturbations, we further utilize the dynamic programming algorithm to select robust features for the matrix construction. The recovered ordinal relationship is then used for contrastive learning on unlabeled samples, and we thus allow more data to be used for feature representation learning, thereby achieving more robust results. The ordinal rankings can also be used to supervise predictions on unlabeled samples, serving as an additional training signal. We provide theoretical guarantees and empirical verification through experiments on various datasets, demonstrating that our method can surpass existing state-of-the-art semi-supervised deep regression methods. Our code have been released on https://github.com/xmed-lab/CLSS.
\end{abstract}


\maketitle
\IEEEdisplaynontitleabstractindextext
\IEEEpeerreviewmaketitle

\section{Introduction}
\label{sec:introduction}

%
%
%
%


 
    \IEEEPARstart{R}{egression} problems are ubiquitous and fundamental in real-world applications, spanning a wide range of tasks and domains including estimating ages from human appearances~\cite{rothe2015dex}, predicting health scores using physiological signals~\cite{engemann2022reusable}, and estimating poses from photographs~\cite{yang2019fsa}. The regression task aims to predict continuous target values, and the most commonly employed approach involves directly predicting the target value and utilizing a distance-based loss function such as $L1$ or $L2$ distance to measure the discrepancy between the prediction and the ground-truth target~\cite{schrumpf2021assessment,zhang2017mpiigaze}. Further, there have been studies tackling regression tasks by employing classification models trained with cross-entropy loss~\cite{niu2016ordinal,rothe2015dex}. By casting regression problems as classification tasks, these methods offer alternative strategies for modeling and optimizing regression models, potentially yielding improved performance and interpretability. Nevertheless, these methods tend to require large amounts of labeled data for training which can be costly to annotate.

    To reduce the reliance on labeled data, semi-supervised learning (SSL), which uses unlabeled data together with a smaller labeled dataset for model training, has been well-studied recently and SSL methods sometimes outperform state-of-the-art (sota) techniques with fully labeled datasets. Chen~\cite{chen2021semi} proposes the semi-supervised CPS for image segmentation, where the mask labeling is time-consuming, and achieves surprising performances. Then UCVME~\cite{li2020transformation} further enforces consistency between transformed inputs and successfully applies the method in both computer vision and medical image regression tasks. To boost regression accuracy, RankUp~\cite{huang2024rankup} instead converts the regression task into the ranking problem and leverages existing semi-supervised classification approaches as the auxiliary ranking classifier. However, these previous methods have primarily concentrated on imposing constraints on the predictions, ignoring explicit regularization imposed on the representations learned by the model. Unfortunately, the learned representations would suffer from fragmentation, lacking the capability of effectively capturing the continuous relationships required in regression tasks~\cite{zha2024rank}.

    Intuitively, contrastive learning provides an alternative technique that effectively improves feature representations in deep neural networks (DNNs), and previous works~\cite{chen2020simple,chen2021empirical,khosla2020supervised,wang2021dense} show that pretraining DNNs through contrastive learning leads to sota classification results. These techniques have since been extended to other vision tasks~\cite{wang2021dense,zhou2005semi,dave2022tclr,radford2021learning,dai2021adaptive,zhang2022improving} to great effect. Further in the SSL setting with abundant unlabeled data, unsupervised contrastive learning has also proven to allow the intrinsic information to be leveraged, benefiting the classification accuracy~\cite{chen2020simple,chen2021empirical,radford2021learning}. However, the learning manner is merely explored for deep regression tasks since labeled data are necessary to ensure feature representations to represent label distance relationships in feature space~\cite{dai2021adaptive,zhang2022improving}. Recently CLSS~\cite{dai2024semi} pioneers contrastive regression in SSL setting with only a small portion of the training dataset labeled. Particularly by enforcing supervised contrastive learning on labeled samples, the remaining unlabeled samples can also be approximately embedded into relatively correct features. Then they employ the spectral seriation algorithm~\cite{atkins1998spectral} to obtain the ordinal relationships between unlabeled samples used for unsupervised contrastive learning. Nevertheless, they ignore the sufficient distance relationship information contained in labeled data which can further rectify the recovered ordinal relationships since they are supervised with ground-truth labels.

    To bridge this gap, in this work, we further extend semi-supervised contrastive regression such that both labeled and unlabeled samples in a mini-batch can be explored and contribute to the accurate recovery of ordinal relationships between unlabeled samples. Specifically in the SSL setting, we observe that the feature similarity matrix constructed with both labeled and unlabeled samples will learn to reflect label distance relationships, where the ground-truth guidance contained in labeled samples further benefits the ordinal ranking recovery and regularizes the regression learning process. Although the feature similarity matrix will be inaccurate and noisy to some extent caused by the instability of unlabeled samples, it is still possible to infer the relative ordering if errors are within certain bounds.
    
\begin{figure*}[h]
    \begin{center}
    \includegraphics[width=0.95\textwidth]{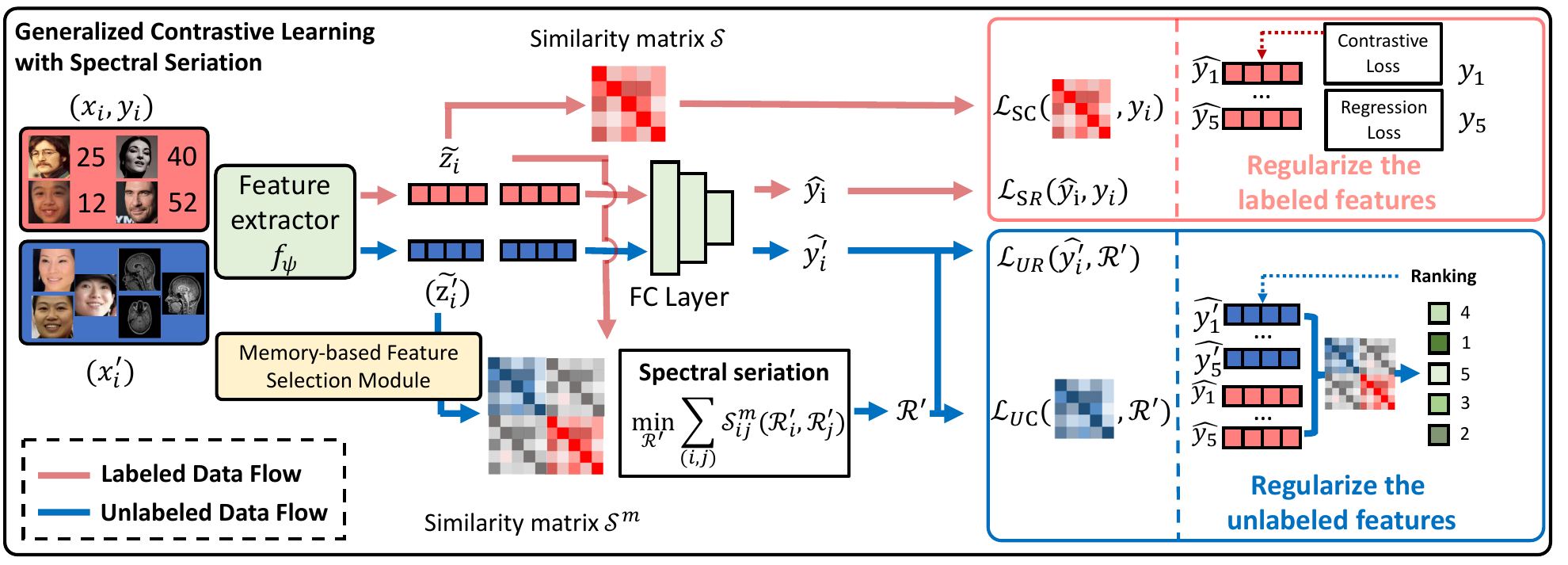}
    \caption{The framework of our proposed GCLSS method, where $({x}_{i}, {y}_{i})$ and $({x}_{i}^{'})$ are labeled pairs and unlabeled samples, respectively, ${\tilde{z}}_{i}$ and ${\tilde{z}}_{i}^{'}$ are the extracted features, and ${\hat{y}}_{i}$ and ${\hat{y}}_{i}^{'}$ correspond to the predictions from labeled and unlabeled samples. Unlike existing contrastive regression works that are only able to employ labeled data to construct supervised regression loss ${\cal{L}}_{SR}$ and supervised contrastive loss ${\cal{L}}_{SC}$, we make use of the spectral seriation to obtain ordinal rankings ${\cal{R}}^{'}$ from unlabeled samples. This can be used for constructing the unsupervised contrastive loss ${\cal{L}}_{UC}$ which serves as contrastive learning and ranking supervision for unlabeled samples.}
    \label{fig:framework}
    \end{center}
    \end{figure*}

    Motivated by this observation, in the following, we propose a novel approach for semi-supervised contrastive regression that leverages ordinal rankings derived from the above-introduced feature similarity matrix. Nevertheless, directly solving with a spectral seriation algorithm (CLSS) cannot give the exact rankings since the introduced labeled features would confound the relative ordinal relationships of unlabeled samples. To accurately extract such ordinal relationships, we extend the spectral seriation algorithm to a generalized version, which allows us to recover rankings of unlabeled samples and construct a distance matrix for guiding contrastive learning on unlabeled samples. The generalized algorithm provides the flexibility to correct errors in extracted rankings from the feature similarity matrix. Further considering the existing perturbations in the feature similarity matrix, we design the memory-based feature selection module (MFSM) to reduce the feature noise and stabilize the learning. Consequently, the obtained ordinal rankings can be effectively utilized to supervise predictions on unlabeled samples, leading to further improvements for semi-supervised regression. We term our method as Generalized Contrastive Learning with Spectral Seriation (GCLSS), as illustrated in Fig.~\ref{fig:framework}.
    
    Moreover, we provide theoretical proofs and analysis, followed by empirically showing that our method can achieve sota performances on multiple datasets. Therefore, our contributions are summarized as follows:
    \begin{itemize}
    \item We propose a novel semi-supervised contrastive regression method, named GCLSS, that explores the relative distance information contained in both labeled and unlabeled data for regression tasks;
    \item To reduce the feature noise and stabilize the training, we design the memory-based feature selection module to provide robust features when constructing the similarity matrix, benefiting the recovery of the ordinal relationships;
    \item We further demonstrate that spectral seriation can be used to extract robust rankings from the feature similarity matrix for supervision on unlabeled samples;
    \item We empirically demonstrate that our method can outperform existing sota alternatives on different deep regression tasks using multiple datasets.
    \end{itemize}
    This paper extends a published work on NeurIPS2023~\cite{dai2024semi} with the following contributions: (1) we extend the CLSS method, which was initially proposed to extract rankings from unlabeled feature similarity matrix for the contrastive regression models while ignoring the ground-truth distance information guidance in labeled samples, with a generalized spectral seriation ranking algorithm. The generalized algorithm further allows labeled samples to provide guidance for ordinal information and thus regularize the ranking solution. The algorithm also provides a closed-form ranking solution even when labeled features are introduced, which confounds the rankings obtained by original spectral seriation algorithm; (2) we also extend the theoretical analysis accordingly with a closed-form solution for the ordinal ranking;  (3) we further design the memory-based feature selection module to lower the feature noises/variations, which reduces the perturbations in the similarity matrix, therefore benefiting the obtained rankings; (4) we accordingly provide similarity matrix noise robustness analysis, confirming that the ordinal ranking is robust to a certain degree even when there exist perturbations/noises in the similarity matrix; (5) we conduct various experiments to demonstrate the effectiveness of GCLSS.


    \section{Related works}
    \subsection{Semi-supervised learning}
    SSL integrates unlabeled data into model training, reducing the need for manual annotations~\cite{dai2022cyclical,li2020transformation,lin2022calibrating,yao2022enhancing,sohn2020fixmatch,zhang2021flexmatch}. Despite the abundance of studies on semi-supervised classification and segmentation, the application of SSL to deep regression task remains less explored~\cite{dai2023semi}. Further investigation is needed to develop techniques that leverage unlabeled data and enhance regression model performances, particularly when labeled data is costly to obtain.

    \subsubsection{Semi-supervised regression} Facing limited labeled distance information, consistency-based methods~\cite{wetzel2022twin,xu2011semi,zhou2005semi} have introduced the utilization of co-trained KNN neural networks to ensure consistency between two models. In recent years, advancements in the field have brought forth iterative techniques~\cite{fazakis2019multi}, deep kernel learning methods~\cite{jean2018semi,mallick2021deep,xu2022semi}, and graph-based approaches~\cite{timilsina2021semi} for semi-supervised regression. Then RankUp~\cite{huang2024rankup} similarly treats the regression task as the ordinal ranking problem, while they constrain the predictions with the designed auxiliary ranking classifier, orthogonal to the contrastive-learning-based CLSS and GCLSS.
    These existing methods, while valuable for certain domains, fall short when it comes to accommodating the complexities and inherent characteristics of unstructured data, such as images and videos. Such data often requires specialized techniques that can effectively capture the rich and high-dimensional feature representations. By addressing this gap, our proposed approach, motivated by unsupervised contrastive learning, aims to extend the benefits of SSL to unstructured inputs, enabling end-to-end training with a deep feature extractor. This advancement allows us to leverage the power of unlabeled data in optimizing the learning for complex and diverse data types, thereby enhancing the performance and adaptability of the model.

    \subsection{Contrastive learning}
    Contrastive learning aims to encompass well-ordered feature representations such that positive and negative sample pairs can be separately grouped and simply recognized~\cite{chen2020simple,he2020momentum,khosla2020supervised,chen2021supercharging}. Depending on how positive pairs are defined, they can be categorized into supervised and unsupervised approaches.
    Supervised contrastive learning methods~\cite{khosla2020supervised} define positive pairs as samples that belong to the same class. In contrast, unsupervised methods~\cite{chen2020simple,he2020momentum}, such as SimCLR, define positive pairs as augmented samples from the same input data, thereby eliminating the need for labels. By leveraging contrastive learning, these techniques facilitate the exploration of relationships between samples and enhance the discriminative power of representations.

    \subsubsection{Contrastive learning for deep regression} 
    In order to accurately recover the label distance relationships of feature representations in the latent space, various contrastive learning approaches have been proposed by researchers. For instance, to encourage similarity values that reflect the pairwise label distance, AdaCon~\cite{dai2021adaptive} employs adaptive margins within the SupCon~\cite{khosla2020supervised} loss function. Then a series of works~\cite{xue2022tractoscr,zhang2022improving,du2024teach} are proposed to further improve the feature representation capability, and they have demonstrated that enhancing feature learning significantly boosts deep regression model performances.
    
    \subsubsection{Contrastive learning for semi-supervised deep regression} 
    To further deal with semi-supervised regression with very limited labeled samples available, the unsupervised contrastive learning method CLSS~\cite{dai2024semi} renders the utilization of unlabeled data during training to regularize the feature representation. They specifically construct the feature similarity matrix with unlabeled features and employ the spectral seriation algorithm to obtain the ordinal relationships between unlabeled samples, which are then used as pseudo-supervision. The learning manner, while proven effective in SSL setting, overlooks the remaining labeled samples. Indeed the feature representations of labeled samples are supervised with ground-truth real-number indicators, which accurately and strictly reflect the label distance relationships. To further utilize the abundant ground-truth information to rectify the recovered ordinal relationships, we take it one step further to construct the feature similarity matrix with both labeled and unlabeled samples in a mini-batch. This exploration enhances the ordinal relationship accuracy when there exist errors in unlabeled features.
    
    \subsection{Feature level variation reduction}
    In the case of semi-supervised regression, unlabeled samples lack the direct guidance of accurate label information, potentially leading to disorganized feature representation. This phenomenon brings challenges when recovering ordinal relationships. To address the problem, uncertainty estimation is commonly used in SSL to adjust for the pseudo-label quality of unlabeled samples. Previous works~\cite{yu2019uncertainty,xia20203d} utilize Monte Carlo dropout to estimate pseudo-label uncertainty, which is then used to filter pseudo-labels and improve the representation stability for segmentation tasks. Further works~\cite{sohn2020fixmatch,zhang2021flexmatch} implicitly filter out pseudo-labels with higher variation by setting confidence thresholds for predictions in semi-supervised classification tasks. Then works~\cite{yao2022enhancing,lin2022calibrating} are proposed to estimate uncertainty based on prediction differences between co-trained models. With similar treatment, UCVME~\cite{dai2023semi} proposes a novel consistency loss that significantly improves model performances and stability in semi-supervised regression tasks.

    These works, while validated in various tasks, directly operate on the image and prediction level, ignoring the constraints on the feature level. Thus the existing variations in unlabeled features lead to unstable training and inaccurate ordinal ranking. To handle the feature level variations, we propose to utilize the dynamic programming algorithm to select features with lower variations in different propagations, therefore enhancing the prediction accuracy.

    \section{Methology}
    In this section, we introduce the improved technique, GCLSS, for the deep regression task. In particular, we briefly review the preliminary knowledge firstly, followed by the proposed generalized contrastive learning with spectral seriation ranking algorithm operated on the feature similarity matrix constructed with both labeled and unlabeled samples. To further stabilize the training process, we additionally introduce the memory-based feature selection module used to reduce the feature variations. Finally, we provide the sensitivity analysis of GCLSS to the perturbations in similarity matrix and features.
    
    \subsection{Preliminary knowledge}
    Before elaborating on the details, we present some necessary notations. The labeled dataset is denoted as ${\mathcal{D}}_{l} = \{ ({x}_{i}, {y}_{i}) \}_{i=1}^{N}$, where $N$ is the total number of labeled samples, and ${x}_{i}$ and ${y}_{i}$ are the input data and corresponding labels, respectively. We denote ${\mathcal{D}}_{ul} = \{ {x}_{i}' \}_{i=1}^{{N'}}$ as the unlabeled dataset consisting of input data only. The feature extractor is denoted as ${f}_{\psi}(\cdot)$ and the feature vector is denoted as $z$ for labeled data, such that ${z}_{i} = {f}_{\psi}({x}_{i})$. Similarly, we denote features for unlabeled data as ${z}_{i}'$. The feature similarity matrix for labeled, unlabeled, and mixed samples (composed of both labeled and unlabeled samples) are denoted as $\mathcal{S}$, $\mathcal{S}'$, and $\mathcal{S}^{m}$, respectively.
    
    Under the above notations, the overall goal of deep-regression is to learn a model $g_{\theta}(y|x)$ capable of predicting the target label $y$ as follows:
    \begin{equation*}
        \hat{y} = {g}_{\theta}({f}_{\psi}(x)) + \epsilon, \quad \epsilon \sim N(0, {\sigma}^{2}),
    \end{equation*}
    where $\hat{y}$ is the predicted label. The conventional supervised regression aims to optimize the model by minimizing mean squared error (MSE) loss, which we denote as:
    \begin{equation}
        {\mathcal{L}}_{SR} = \frac{1}{N} \sum_{i=1}^{N}{({y}_{i} - {\hat{y}}_{i})}^{2}.
    \end{equation}
    In order to improve the discriminability, contrastive learning is performed on $L2$ normalized feature vectors, with some methods also performing feature projection onto a lower dimension~\cite{chen2020simple,zhang2023improving}. We denote the transformed, normalized feature vector for labeled and unlabeled data as ${\widetilde{z}}_{i} = \frac{{z}_{i}}{\|{z}_{i}\|_{2}} $ and ${\widetilde{z}}_{i}'=\frac{{z}_{i}'}{\|{z}_{i}'\|_{2}}$, respectively.
	
    \subsection{Using ordinal ranking from spectral seriation for supervision}
    Although contrastive learning supervision ${\mathcal{L}}_{SC}$ discriminates features corresponding to different labels, it overlooks the intrinsic ordinal relation between unlabeled features. Fortunately, feature similarity matrix provides measurement between involved samples, and spectral seriation is a proven procedure to recover the ordinal rankings from the matrix. In this manner, we are capable of extracting the ordinal rankings of unlabeled samples $\mathcal{R}'$ from $\mathcal{S}$, $\mathcal{S}'$, and $\mathcal{S}^{m}$, which are then used to supervise contrastive learning and predictions on unlabeled samples. In the following, we provide a detailed description of our method.
    
    \subsubsection{Spectral seriation for retrieving ordinal ranking}
    Basically, contrastive regression focuses on ensuring that feature representations accurately capture the distance relationships between labels. That is to say, if ${y}_{i}$ and ${y}_{j}$ are closer to each other than ${y}_{i}$ and ${y}_{k}$, then the distance between the corresponding features ${z}_{i}$ and ${z}_{j}$ should be smaller than the distance between ${z}_{i}$ and ${z}_{k}$. In other words:
    \begin{equation*}
        \| \widetilde{{z}}_{i} - \widetilde{{z}}_{j} \|_{2} < \| \widetilde{{z}}_{i} - \widetilde{{z}}_{k} \|_{2} \quad\mbox{for}\quad | {y}_{i} - {y}_{j} | < |{y}_{i} - {y}_{k} |.
    \end{equation*}
    When substituting the $L2$ norm with the cosine similarity function $\gamma$, we obtain:
    \begin{equation*}
        \gamma(\widetilde{{z}}_{i}, \widetilde{{z}}_{j}) > \gamma(\widetilde{{z}}_{i}, \widetilde{{z}}_{k}) \quad\mbox{for}\quad | {y}_{i} - {y}_{j} | < |{y}_{i} - {y}_{k} |.
    \end{equation*}
    Intuitively, $\gamma(\widetilde{{z}}_{i}, \widetilde{{z}}_{j})$ and $\gamma(\widetilde{{z}}_{i}, \widetilde{{z}}_{k})$ correspond to entries in the similarity matrix $\mathcal{S}$.

    To obtain the ordinal ranking $\mathcal{R}'$ of samples in a batch of unlabeled data, we can leverage the spectral seriation algorithm initially introduced in~\cite{atkins1998spectral}. This algorithm is specifically designed to recover an ordinal ranking based on a correlation matrix, where higher correlation values indicate a closer proximity in terms of ranking. The spectral seriation algorithm is particularly beneficial when determining the similarity between pairs of samples is straightforward, but establishing a direct ordering is challenging. Since cosine similarity is equivalent to correlation after $L2$ normalization, we can utilize spectral seriation to recover $\mathcal{R}'$ from $\mathcal{S}'$. This allows us to effectively capture the ordinal relationships within the data.

    Specifically, the seriation problem can be formulated as the following objective:
    \begin{equation}
    \label{pro:seriation}
        {\mathcal{R}}'^{*} = \arg\min_{\mathcal{R}'} \sum_{i,j} {\mathcal{S}}_{[i,j]}' ({\mathcal{R}}_{i}' - {\mathcal{R}}_{j}')^{2},
    \end{equation}
    where ${\mathcal{S}}_{[i,j]}' = \gamma(\widetilde{{z}}_{i}', \widetilde{{z}}_{j}')$ is the feature similarity matrix only containing unlabeled samples. Since sample pairs closer together in ranking have higher correlation values, minimizing this objective encourages these samples to have ${\mathcal{R}}_{i}'$ and ${\mathcal{R}}_{j}'$ that are closer together. As per spectral seriation, the solution to $\mathcal{R}'$ can be derived from the Fiedler vector, as stated in the following Theorem.

    \begin{theorem}
        \label{theo1}
        Given similarity matrix $\mathcal{S}'$, such that ${\mathcal{S}}_{[i,j]}' > {\mathcal{S}}_{[i,k]}'$ for $| {y}_{i} - {y}_{j} | < |{y}_{i} - {y}_{k} |$, the ordinal ranking that best satisfies the observed $\mathcal{S}'$ is the ranking of the values in the Fiedler vector of $\mathbf{L}'$, where $\mathbf{L}'$ is the Laplacian of $\mathcal{S}'$.
    \end{theorem}
    The proof of the theorem can be directly obtained by approximating discrete rankings $R'$ with real-number values $r'$ and expressing Eq.~\ref{pro:seriation} in the form of:
    \begin{equation*}
        \min_{{r'}^{\top}e=0, {r'}^{\top}r'=1} {r'}^{\top}\mathbf{L}r',
    \end{equation*}
    where $r'$ can be solved by computing the Fiedler vector, which is the eigenvector corresponding to the smallest non-zero eigenvalue. The relative rankings of $r'$ then give us the estimation of $\mathcal{R}'$. 
    
    \subsubsection{Generalized seriation rankings with mixed samples}
    Based on the above-introduced theorem and the obtained ranking solution $\mathcal{R}'$ of a batch of unlabeled samples for each anchor sample ${x}_{i}'$, CLSS~\cite{dai2024semi} defines the unlabeled contrastive loss with respect to the subset $\mathcal{B}$ as:
    \begin{equation}
        \label{clss}
        \sum_{i=1}^{| \mathcal{B} |} \mathcal{L}( \mathbf{rk}({\mathcal{S}}_{[i,: ]}'), \mathbf{rk}(-| \mathcal{R}' - {\mathcal{R}}_{i}' |); 
    \end{equation}
    where $\mathbf{rk}$ denotes the ranking operator, $[i,:]$ denotes the $i$-th row of the matrix, and $\mathcal{L}$ denotes the ranking similarity function. Although recovering the ordinal ranking solely from the unlabeled batch seems effective in CLSS~\cite{dai2024semi}, the unlabeled features tend to be noisy and will be mapped to incorrect locations in the feature space with small perturbations, leading to the inaccurate rankings. In contrast, labeled features are guided by ground-truth labels, making the features reliable and represent accurate ordinal information. To further utilize the ignored information contained in labeled samples, we generalize Theorem~\ref{theo1} as follows:

    \begin{corollary}
    \label{gclss}
        Given the well-ordered similarity matrix $
        {\mathcal{S}}^{m} \in \mathbb{R}^{(m+n) \times (m+n)}$ of a batch of $m$ labeled and $n$ unlabeled samples, such that ${\mathcal{S}}_{[i,j]}^{m} > {\mathcal{S}}_{[i,k]}^{m}$ for $| {y}_{i} - {y}_{j} | < |{y}_{i} - {y}_{k} |$. Then the ordinal ranking of unlabeled samples that best satisfies ${\mathcal{S}}^{m}$ is the ranking of values in the sub-Fiedler vector of $\mathbf{L}^{m}\triangleq \begin{pmatrix} 
        \mathbf{L} & {\mathbf{L}}_{m} \\ 
        {\mathbf{L}}_{m}  & \mathbf{L}'  
        \end{pmatrix}$, which is the Laplacian of $\mathcal{S}^{m}$, and the sub-Fiedler vector of $\mathbf{L}^{m}$ has the following closed-form:
        \begin{equation}
            (\sum_{i=1}^{n}\frac{1}{{\lambda}_{i}}{\alpha}_{i}{\alpha}_{i}^{\top})(-{\mathbf{L}}_{m}r),
        \end{equation}
        where $\{{\lambda}_{i}\}_{i=i}^{n}$ are eigenvalues of $\mathbf{L}'$ with the non-descending order, $\{{\alpha}_{i}\}_{i=i}^{n}$ are corresponding eigenvectors with $\| {\alpha}_{i} \|=1$, and $r$ can be solved via computing the Fiedler vector of $\mathbf{L}$ .
    \end{corollary}

    Particularly the proof is similar to the Theorem.~\ref{theo1}, and can be obtained by solving the following objective:
    \begin{equation}
        \min_{r'} (r^{\top}, r'^{\top}) \mathbf{L}^{m}
        \begin{pmatrix} 
        r \\ 
        r' 
        \end{pmatrix},
    \end{equation}
    where $\mathbf{L}^{m}$ is the Laplacian of the well-ordered ${\mathcal{S}}^{m}$. The specific proof is provided in the appendix.
    
    \subsubsection{Ranking supervision of unlabeled predictions with generalized seriation rankings}
    With the ground-truth label distance information, the generalized ranking obtained from spectral seriation algorithm gives us the more accurate $\mathcal{R}'$ of the observed batch of unlabeled samples. In this manner, we can still obtain rankings for each anchor sample ${x}_{i}'$. Thus similar to the loss function~\ref{clss} in CLSS, we further define the unlabeled contrastive learning loss with respect to some subset $\mathcal{B}$ as:
    \begin{equation*}
        {\mathcal{L}}_{UC} = \sum_{i=1}^{| \mathcal{B} |} \mathcal{L}( \mathbf{rk}({\mathcal{S}}_{[i,: ]}'), \mathbf{rk}(-| \mathcal{R}' - {\mathcal{R}}_{i}' |); 
        \lambda),
    \end{equation*}
    where $\mathcal{R}'$ is obtained via Corollary~\ref{gclss} and regularized by the ground-truth guidance from labeled samples. This loss function ensures that the ranking of feature similarity values with some anchor sample ${x}_{i}'$ are consistent with that of the distances between derived rankings obtained from seriation. Note that for the differentiable ranking similarity loss function $\mathcal{L}$, we directly use the differential combinatorial solver proposed in~\cite{poganvcic2019differentiation}, and the parameter $\lambda$ is additionally introduced to  control the step.

    \begin{algorithm}[b!]
	\caption{Dynamic Programming Algorithm}
	\label{alg1}
	\LinesNumbered 
	\KwIn{${\mathcal{V}}$, the $|\mathcal{B}|\times |\mathcal{B}|$ variation matrix of feature similarity values, the pre-defined number $B'$ of totally selected features.}
	\KwOut{$I_{[|\mathcal{B}|, B']} - 1$.}
	Initialize the selection matrix $I\in \mathbb{R}^{(|\mathcal{B}|+1)\times (B' + 1)}$ with $(|\mathcal{B}|+1)\times (B' + 1)$ empty lists\;
        Initialize the cost matrix $D\in \mathbb{R}^{(|\mathcal{B}|+1)\times (B' + 1)}$ with $(|\mathcal{B}|+1)\times (B' + 1)$ infinity items\;
        Set the first two columns of $D$ as $0$, $D_{[0:2,:]} \leftarrow 0 $ \;
	\For{${n}_{1} = 3 : |\mathcal{B}|+1 $ }{ 
            \For{$b = 3 : \min\{{n}_{1}, B'\} + 1$}
            {
                \eIf{$b = 0$}{
		          $D_{[{n}_{1}, b]} = 0$\;
                    $I_{[{n}_{1}, b]} = [$ $]$\;
		          }{
		          $D_{[{n}_{1}, b]} = D_{[{n}_{1} - 1, b]}$\;
                    $I_{[{n}_{1}, b]} = I_{[{n}_{1} - 1, b]}$\;
                    \For{${n}_{2} = 0 : {n}_{1}$}
                    {
                        \If{$D_{[{n}_{1}, b]} < D_{[{n}_{2},b-1]} + \sum_{j\in I_{[{n}_{2},b-1]}}\mathcal{V}_{[{n}_{1},j]} $}
                        {
                        $D_{[{n}_{1}, b]} = D_{[{n}_{2},b-1]} + \sum_{j\in I_{[{n}_{2},b-1]}}\mathcal{V}_{[{n}_{1},j]}$\;
                        $I_{[{n}_{1}, b]} = I_{[{n}_{2}, b - 1]}.\text{append}({n}_{1}) $\;
                        }
                    }
		}
		}
	}
        return
    \end{algorithm}

    After obtaining the ranking $\mathcal{R}'$ for unlabeled samples, we can also perform supervision on the prediction $\hat{y}$. Since spectral seriation algorithm is inherently robust to errors, which will be shown later, the pair-wise distance rankings from seriation are likely to be more accurate than the predicted output. Therefore we define the unlabeled prediction ranking loss with respect to some subset $\mathcal{B}$ as:
    \begin{equation*}
        {\mathcal{L}}_{UR} = \sum_{i=1}^{| \mathcal{B} |} \mathcal{L}( \mathbf{rk}(-| \hat{y} - {\hat{y}}_{i} |), \mathbf{rk}(-| \mathcal{R}' - {\mathcal{R}}_{i}' |); 
        \lambda).
    \end{equation*}
    Note that the loss function serves as an additional regularization supervision for unlabeled data. Thus, the overall loss function of our method is 
    \begin{equation*}
        {\mathcal{L}}_{full} = {\mathcal{L}}_{SR} + {\lambda}_{SC}{\mathcal{L}}_{SC} + {\lambda}_{UC}{\mathcal{L}}_{UC} + {\lambda}_{UR}{\mathcal{L}}_{UR},
    \end{equation*}
    where ${\mathcal{L}}_{SR}$, ${\mathcal{L}}_{SC}$, ${\mathcal{L}}_{UC}$, and ${\mathcal{L}}_{UR}$ are loss functions for supervised regression, supervised contrastive learning, unsupervised contrastive learning, and unsupervised ranking, respectively. The hyper-parameters ${\lambda}_{SC}$, ${\lambda}_{UC}$, and ${\lambda}_{UR}$ are introduced balancing parameters.
 
    \subsection{Memory-based feature selection module}
    Different from classification task, which can smooth predictions using thresholding functions for class pseudo-labeling, deep-regression tasks directly use real number predictions as pseudo-labels. Therefore, model performance highly depends on the quality of pseudo-labels, i.e. predictions.     
    Particularly when solving the above seriation-based rankings, which provides pseudo-supervision for the unlabeled samples and operates on the feature level, the feature quality therefore determines the pseudo-supervision quality. Thus, to ensure a stable learning process, one of the key point is to keep features accurate, i.e. with lower noise.

    \begin{table}[t!]
    \begin{center}
    \caption{The experimental results of DP algorithm on the toy example. Specifically ``GT" and ``DP" represents the ground-truth label and prediction label of samples with the smallest variance, respectively. In line with expectations, DP algorithm achieves $50\%-70\%$ accuracy, i.e., successfully selecting $50\%-70\%$ samples with the smallest noise (variances).}
    \label{table:toy-example}
    \begin{tabular}{c|c|c}
    \hline
           &Labels  & Accuracy \\ 
    \hline
    GT     &[1, 2, 3, 6, 8, 10, 11, 13, 14, 22]  &\multirow{2}{*}{70\%}  \\ 
    DP     &[0, \textbf{1}, \textbf{2}, \textbf{8}, \textbf{10}, \textbf{11}, \textbf{13}, 21, \textbf{22}, 24]  &  \\ 
    \hline
    GT     &[1, 3, 4, 9, 12, 16, 17, 19, 20, 23]   &\multirow{2}{*}{60\%}  \\ 
    DP     &[0, \textbf{1}, 2, \textbf{3}, \textbf{9}, 10, \textbf{17}, \textbf{20}, 21, \textbf{23}]&  \\ 
    \hline
    GT     &[2, 4, 8, 10, 14, 16, 17, 19, 21, 24]  &\multirow{2}{*}{60\%}  \\ 
    DP     &[\textbf{2}, 3, 6, \textbf{8}, \textbf{10}, 12, \textbf{16}, 18, \textbf{21}, \textbf{24}]  &  \\
    \hline
    GT     &[1, 4, 9, 13, 14, 15, 19, 20, 22, 24] &\multirow{2}{*}{50\%}  \\ 
    DP     &[\textbf{1}, 2, 6, 10, 11, \textbf{15}, 18, \textbf{19}, \textbf{22}, \textbf{24}]    &  \\
    \hline
    GT     &[1, 2, 3, 7, 10, 11, 14, 18, 19, 21]    &\multirow{2}{*}{70\%}  \\ 
    DP     &[\textbf{1}, \textbf{2}, \textbf{3}, 4, 5, \textbf{7}, \textbf{10}, \textbf{14}, \textbf{19}, 22]   &  \\
    \hline
    \end{tabular}
    \end{center}
    \end{table}
    
    \subsubsection{Memory-based feature selection module} To reduce the affiliated noise and stabilize the training, we adpot a memory-based feature selection module for GCLSS, consisting of two propagation process. The first forward propagation provides a group of features $\{{z}_{i1}'\}_{i=1}^{|\mathcal{B}|}$ of the batch of unlabeled samples, and the back propagation is employed to finetune the model parameters, leading to the other group of features $\{{z}_{i2}'\}_{i=1}^{|\mathcal{B}|}$ of the same batch of samples. In this manner, four similarity matrices $\{{\mathcal{S}}^{11}, {\mathcal{S}}^{12}, {\mathcal{S}}^{12}, {\mathcal{S}}^{22}\}$ constructed by the two groups of features can be obtained. Since ${z}_{i1}'$ and ${z}_{i2}'$ are indeed two representations of the same sample, each entry $\{{\mathcal{S}}^{mn}_{[i,j]}\}_{m,n=1}^{2}$ of these four feature similarity matrices provides the measurement between features of the two groups. \textit{Intuitively, entries with lower variation point out that the features contain less noise and are more robust to the representation, therefore more accurate and benefiting the training.} Motivated by the observation, we firstly compute the variation matrix:
    \begin{equation*}
        \begin{split}
            \mathcal{V} &= \text{Var}({\mathcal{S}}^{11}, {\mathcal{S}}^{12}, {\mathcal{S}}^{12}, {\mathcal{S}}^{22}), \\
        \end{split}
    \end{equation*}
    where Var is the operator to calculate the variation of each entry among the four matrices.
    Then, with the dynamic programming algorithm as described in Algorithm~\ref{alg1}, we select the final candidate features with lower noise:
    \begin{equation}
    \begin{split}
        \mathcal{Z'} &= \text{DP}(\mathcal{V}; B'), \\
    \end{split}
    \end{equation}
    where $B'$ is a pre-defined number of totally selected features, and $\mathcal{Z'} $ is the selected features in the batch. These selected features are then employed to compute the above ranking $R'$, the loss functions ${\mathcal{L}}_{UC} $, and $ {\mathcal{L}}_{UR}$.
    
    \subsubsection{Toy experiment verification}
    To evaluate if our proposed DP algorithm can effectively select features with lower noise, we pre-conduct experiments as a toy example. Specifically we randomly generate $25$ vectors $\{{z}_{i}^{t}\}_{i=1}^{25} \in {\mathbb{R}}^{256}$ as the simulated feautures in the low-dimensional latent space. With repeating the process $10$ times, we obtain $10$ groups of features $\{{z}_{ij}^{t}\}_{i,j=1}^{25,10}$ which are regarded as $10$ different observations of the batch of samples. Then we explicitly add noises $\{{n}_{ij}\}_{i,j=1}^{25,10}$ to these simulated features where these noises $\{{n}_{ij}\}_{i=1}^{25}$ follow pre-defined Guassian distributions $\{\mathcal{N}_{i}(0,{\sigma}_{i})\}_{i=1}^{25}$, respectively. Higher variances indicate higher perturbations, therefore noisier features. With these preprocessing, we obtain the involved noisy features. Next, we compute the feature similarity matrices $\{{\mathcal{S}}^{ij}\}_{i,j=1}^{10}$, followed by employing our DP algorithm to select features with lower noise (variance). For the hyper-parameters, we set $B'$ as $10$, i.e., selecting $10$ features from the $25$ ones, and results are shown in Table.~\ref{table:toy-example}. To avoid contingency, we repeat the process for $5$ times, and DP algorithm achieves $62\%$ accuracy in average, i.e., successfully selecting $50\%-70\%$ examples with the smallest noise (variance). The selection accuracy meets the requirement to a certain extent and verifies the effectiveness of our memory-based selection module.

    
    \subsection{Robustness analysis of spectral seriation}
    Based on the loss minimization process, spectral seriation is inherently robust to noisy inputs. Next we present theoretical proofs to formally establish the robustness of the generalized seriation rankings against two distinct types of noise, i.e., the noisy similarity matrices and noisy feature representations. Via analytical derivations, we provide approximate bounds to demonstrate that the generalized seriation rankings can reliably recover ordinal rankings even when faced with noisy similarity values in matrix ${\mathcal{S}}^{m}$. For readers seeking more technical details, we have included additional derivations in the appendix.

    The main result is presented in Theorem~\ref{theo2}. To build up the proof, we firstly present two related lemmas , where Lemma~\ref{lemma:1} provides the perturbation bounds for eigenvalues of symmetric matrices, and Lemma~\ref{lemma:2} provides the upper bounds of the eigenvectors.

    \begin{lemma}
    \label{lemma:1}
    Let $\lambda_i$ be an eigenvalue of the Laplacian matrix $\mathbf{L}$ of some similarity matrix and the corresponding normalized eigenvector is $\alpha_i$. Suppose that $\Delta\mathbf{L}$ is the perturbation of $\mathbf{L}$ and $\Delta \lambda_i + \lambda_i$ and $\Delta \alpha_i + \alpha_i$ are the perturbed eigenvalue and eigenvector of $\mathbf{L} + \Delta \mathbf{L}$. Then we have:
    \begin{align}
    \Delta \lambda_i = \alpha_i^\top \Delta \mathbf{L} \alpha_i+O(\|\Delta \mathbf{L}\|_2^2).
    \end{align}
    \end{lemma}

    \begin{lemma}
    \label{lemma:2}
    Let $\lambda_{i}$ and $\alpha_{i}$ be the eigenvalue and corresponding eigenvector of the Laplacian matrix $\mathbf{L}$ of the symmetric matrix ${\mathcal{S}}^{m}$. The corresponding perturbations are denoted as $\Delta \lambda_i, \Delta \alpha_i, \text{ and } \Delta \mathbf{L}$. Let $\lambda_{1} < \lambda_{2} < \dots < \lambda_{n}$, then we have
    \begin{align}
    \Delta \alpha_i = \sum_{j\neq i}\frac{\alpha_{j}^\top \Delta \mathbf{L} \alpha_i}{\lambda_i-\lambda_j}\alpha_{j}+O(\|\Delta \mathbf{L}\|_2^2).
    \end{align}
    \end{lemma}
    We next present the main theorem. 
    \begin{theorem}
    \label{theo2}
    Consider the feature similarity matrix ${\mathcal{S}}^{m} \in \mathbb{R}^{(m+n) \times (m+n)}$ for $m$ labeled and $n$ unlabeled samples. 
    Assume the corresponding perturbation is $\Delta \mathcal{S}^{m}$. When 
    \begin{align}
    \label{eq:deltaS}
    \|\Delta \mathcal{S}^{m}\|_{\infty} \leq \min\{\frac{\sigma_1(\mathbf{L}_{m})}{2}, \frac{ \lambda_1^2\sigma_{1}(\mathbf{L}_{m}){\min}_{i,j,k}|\alpha_{ik}-\alpha_{jk}|}{8 \sqrt{2}\lambda_n n\|\mathbf{L}_{m}\|_{\infty}}\},
    \end{align}
    the generalized seriation rankings obtained with ${\mathcal{S}}^{m}$ is the same as that obtained using ${\mathcal{S}}^{m}+\Delta{\mathcal{S}}^{m}$, indicating that our algorithm is robust to noise in $\mathcal{S}^{m}$. Note that $\sigma_1({\mathbf{L}}_{m})$ denotes the minimum singular value of ${\mathbf{L}}_{m}$, $\{{\lambda}_{i}\}_{i=1}^{n}$ are the eigenvalues of ${\mathbf{L}}_{m}$ with non-descending order, $\{{\alpha}_{i}\}_{i=1}^{n}$ are the corresponding eigenvectors, and ${\alpha}_{ik}$ is the $k$-th value of eigenvector ${\alpha}_{i}$.
    \end{theorem}
    Note that the proof of Theorem~\ref{theo2} relies mainly on the definition of the Fielder vector and the previous lemmas. Detailed derivations are included in the appendix.
    
    Then, if we consider the noises in the learned feature representations, in fact, this is a special case of the above theorem. We present the main findings in the following corollary.
    \begin{corollary}
    For the similarity matrix ${\mathcal{S}}^{m} \in \mathbb{R}^{(m+n) \times(m+n)}$, suppose the row $m+i$ and column $m+i$ are corrupted due to inadequate feature representations ${\widetilde{z}}_{i}'$ being learned for unlabeled sample ${x}_{i}'$. When 
    \begin{align}
    \label{eq:deltaz}
    ||\Delta {\widetilde{z}}_{i}'||_{2} \leq \frac{\min\{\frac{\sigma_1(\mathbf{L}_{m})}{2}, \frac{ \lambda_1^2\sigma_{1}(\mathbf{L}_{m}){\min}_{i,j,k}|\alpha_{ik}-\alpha_{jk}|}{8 \sqrt{2}\lambda_n n\|\mathbf{L}_{m}\|_{\infty}}\}}{m+n-1},
    \end{align}
    the generalized seriation rankings obtained by the spectral seriation algorithm is stable, where the proof is similar to the above theorem.
    \end{corollary}

    \section{Experimental results}
    We next evaluate our proposed method on four datasets to verify the effectiveness for semi-supervised deep regression task. In order to include diverse image types, we conduct experiments on IXI dataset, the medical imaging dataset for brain age estimation from MRI volumes, a synthetic non-linear dataset for operator learning, and two natural image datasets for age estimation from photography. For clearly understanding these tasks, we have selected samples and visualize them in Fig.~\ref{fig:samples}. Particularly in experiments, we will first conduct the ablation studies on IXI dataset to select the best suitable hyper-parameters and exhibit the comparisons with sota methods on the dataset, followed by verifying the effectiveness through comparisons on the other datasets. Note that we utilize PyTorch framework and train models on the Nvidia RTX3090 GPU for implementation.
    
    \subsection{Validation on brain age estimation from MRIs}
    Semi-supervised deep regression problems hold significant value, particularly in medical applications where real-number medical indicators are prevalent for disease tracking~\cite{jean2018semi,ouyang2020video}. Nevertheless, acquiring labeled data from patients can be challenging due to privacy concerns and associated costs~\cite{dai2022cyclical,li2020transformation}. In this scenario, our proposed method proves to be particularly beneficial.
    Specifically, to validate the effectiveness of our GCLSS, we conduct experiments using the IXI brain MRI dataset for brain age estimation task~\cite{franke2010estimating}. The task particularly involves training a model to extract relevant phenotypes from MRIs that are indicative of brain health, offering crucial applications in disease detection~\cite{cole2017predicting,gaser2013brainage}. Subsequently, the trained model is employed to estimate the biological brain age of unhealthy patients, facilitating disease screening~\cite{armanious2021age}.

    \begin{figure}[t]
    \begin{center}
    \includegraphics[width=0.5\textwidth]{images.pdf}
    \caption{Visualizations of sample data from IXI, AgeDB-DIR, and UTKFace datasets for age estimation. Images are shown with paired age labels, which is used for training.}
    \label{fig:samples}
    \end{center}
    \end{figure}

    \begin{table*}[t!]
    \setlength\tabcolsep{3pt}
    \begin{center}
    \caption{The MAE and $\mathbf{R}^2$ results on IXI dataset. GCLSS consistently performs the best, and boosts the performances of CLSS with the regularization information from labeled samples, which confirms the effectiveness of the ranking solution based on ${\mathcal{S}}^{m}$.}
    \label{tab:ixi}
    \resizebox{\textwidth}{!}{
    \begin{tabular}{  c l | l | cccc | cccc }
    \cline{1-11}
    \multirow{9}{*}{\rotatebox{90}{IXI dataset}}&\multirow{2}{*}{Type}&\multirow{2}{*}{Method}&\multicolumn{4}{c|}{MAE $\downarrow$}&\multicolumn{4}{c}{$\mathbf{R}^2$$\uparrow$}    \\ 
    \cline{4-11}
    &&&1/5labels&1/4labels&1/3labels&1/2labels &1/5labels&1/4labels&1/3labels&1/2labels \\
    \cline{2-11}
    &\textit{Supervised} &Regression&9.95$\pm$1.41&11.93$\pm$1.40&11.76$\pm$1.75&10.93$\pm$1.60  & 43.1\%$ \pm$14.4 & 20.2\%$ \pm$16.3 & 23.3\%$ \pm$20.7 & 33.3\% $\pm$16.7 \\ 
    \cline{2-11}
    &&Mean-Teacher~\cite{tarvainen2017mean}&11.23$\pm$2.31&10.27$\pm$1.57&10.52$\pm$3.12&12.01$\pm$2.03  &39.1\%$ \pm$21.3 & 22.9\%$ \pm$19.3 & 23.4\%$ \pm$25.3 & 36.3\%$ \pm$18.5 \\ 
    \cline{3-11}
    &\textit{Semi-}&CPS~\cite{chen2021semi}&10.23$\pm$1.41&10.27$\pm$1.19&\underline{9.64$\pm$1.27}&9.69$\pm$1.01 &39.0\%$\pm$14.2 & 39.4\%$\pm$13.0 & \underline{46.2\%$\pm$12.7} & 46.2\%$\pm$10.6 \\ 
    \cline{3-11}
    &\textit{Supervised}&UCVME~\cite{dai2023semi}&9.83$\pm$1.32&10.86$\pm$1.67&9.65$\pm$1.31&10.06$\pm$1.19 &41.7\%$\pm$13.1 & 32.3\%$\pm$17.7 & 45.5\%$\pm$12.6 & 39.9\%$\pm$12.1 \\
    \cline{3-11}
    &&CLSS~\cite{dai2024semi}&\underline{9.58$\pm$1.48}&\underline{9.68$\pm$1.22}&9.72$\pm$1.29&\underline{9.37$\pm$1.17}&\underline{45.0\%$\pm$17.6}&\underline{44.5\%$\pm$11.5}&44.9\%$\pm$14.9&\underline{48.9\%$\pm$13.2} \\
    \cline{3-11}
    \cline{3-11}
    &&GCLSS&\textbf{8.98 $\pm$ 0.34}&\textbf{9.01$\pm$0.72}&\textbf{9.33$\pm$0.92}&\textbf{8.82$\pm$0.49} &\textbf{53.3\%$\pm$3.0}&\textbf{49.3\%$\pm$7.0}&\textbf{48.9\%$\pm$7.9}&\textbf{51.0\%$\pm$3.7} \\
    \cline{1-11}
    \end{tabular}
    }
    \end{center}
    \end{table*}

    \begin{table*}
    \setlength\tabcolsep{3pt}
    \begin{center}
    \caption{The technique ablation results on IXI dataset. The abbreviation uldr, ldr and MFSM represents ``unlabeled data for ranking", ``labeled data for ranking", and ``memory-based feature selection module", respectively. Results show that both the proposed generalized seriation ranking and feature selection module contribute to the final results.}
    \label{tab:ablation-module}
    \begin{tabular}{  l| ccc | cccc | cccc }
    \hline
    \multirow{2}{*}{Method} & \multirow{2}{*}{uldr} & \multirow{2}{*}{ldr} & \multirow{2}{*}{MFSM} &\multicolumn{4}{c}{MAE $\downarrow$} &\multicolumn{4}{|c}{$\mathbf{R}^2$$\uparrow$}\\
    \cline{5-12}
    &&&&1/5labels&1/4labels &1/3labels &1/2labels &1/5labels&1/4labels &1/3labels &1/2labels\\
    \hline
    CLSS&\checkmark&&&9.58$\pm$1.48&9.68$\pm$1.22&9.72$\pm$1.29&9.37$\pm$1.17&45.0\%$\pm$17.6&44.5\%$\pm$11.5&44.9\%$\pm$14.9&48.9\%$\pm$13.2 \\
    CLSS+MFSM&\checkmark&&\checkmark&9.49$\pm$1.02&9.11$\pm$0.69&9.64$\pm$0.78&9.23$\pm$0.62&43.9\%$\pm$10.3&48.5\%$\pm$7.0 &44.6\%$\pm$8.6 &48.6\%$\pm$4.9 \\
    CLSS+ldr&\checkmark&\checkmark&&9.50$\pm$1.70&9.39$\pm$1.31&9.84$\pm$1.48&8.85$\pm$0.55&42.6\%$\pm$18.4&46.5\%$\pm$13.5&42.2\%$\pm$15.6&52.9\%$\pm$4.2 \\
    GCLSS&\checkmark&\checkmark&\checkmark&\textbf{8.98 $\pm$ 0.34}&\textbf{9.01$\pm$0.72}&\textbf{9.33$\pm$0.92}&\textbf{8.82$\pm$0.49}&\textbf{53.3\%$\pm$3.0}&\textbf{49.3\%$\pm$7.0}&\textbf{48.9\%$\pm$7.9}&\textbf{51.0\%$\pm$3.7} \\
    \hline
    \end{tabular}
    \end{center}
    \end{table*}

    \begin{table}[t!]
    \begin{center}
    \caption{Ablation results using different batch sizes for unlabeled data sampling on IXI dataset, and the best result is obtained when setting to $8$.}
    \label{table:batchsize-ixi}
    \begin{tabular}{c|c|c}
    \hline
    Batch Size & MAE $\downarrow$ & $\mathbf{R}^2$$\uparrow$ \\ 
    \hline
    4   & 10.17 $ \pm$ 1.88 & 38.0\% $ \pm$ 19.7 \\
    8   & \textbf{9.05$\pm$0.25} & \textbf{52.6\%$\pm$2.7} \\
    16  & 9.33 $ \pm$ 1.47 & 46.7\% $ \pm$15.1 \\
    32   & 12.37 $ \pm$ 1.34 & 17.9\% $ \pm$ 13.3 \\
    \hline
    \end{tabular}
    \end{center}
    \end{table}

    \begin{table}[t!]
    \begin{center}
    \caption{Ablation results using different hyper-parameter values on IXI dataset, and our model performs the best when ${\lambda}_{SC} = 0.1$, ${\lambda}_{UC}=0.05$ and ${\lambda}_{UR}=0.01$, which are fixed in later experiments.}
    \label{table:hypers-ixi}
    \begin{tabular}{c|c|c|c}
    \hline
    Weights &Value & MAE $\downarrow$ & $\mathbf{R}^2$$\uparrow$ \\ 
    \hline
                       & 1.0   & 9.31$\pm$0.56 & 47.0\%$\pm$6.6 \\
    ${\lambda}_{SC}$   & 0.1   & \textbf{9.05$\pm$0.25} & \textbf{52.6\%$\pm$2.7} \\
                       & 0.01  & 9.18$\pm$0.62 & 50.2\%$\pm$6.9 \\
    \hline
                       & 0.5   & 9.88$\pm$1.37 & 43.2\%$\pm$12.9 \\
    ${\lambda}_{UC}$   & 0.05  & \textbf{9.05$\pm$0.25} & \textbf{52.6\%$\pm$2.7} \\
                       & 0.005 & 9.41$\pm$1.09 & 47.4\%$\pm$11.9 \\
    \hline
                       & 0.1   & 9.46$\pm$0.75 & 47.9\%$\pm$6.3 \\
    ${\lambda}_{UR}$   & 0.01  & \textbf{9.05$\pm$0.25} & \textbf{52.6\%$\pm$2.7} \\
                       & 0.001 & 10.06$\pm$1.07 & 41.1\%$\pm$11.3 \\
    \hline
    \end{tabular}
    \end{center}
    \end{table}

    \begin{table}[h!]
    \begin{center}
    \caption{Ablation results using different $B'$ values on IXI dataset, and GCLSS performs the best when $B'=6$. Larger ones bring larger variations and small ones are with limited relative distance information.}
    \label{table:B'}
    \begin{tabular}{l|cc}
    \hline
    $B'$ & MAE $\downarrow$ & $\mathbf{R}^2$$\uparrow$ \\ 
    \hline
    8 & 9.05$\pm$ 0.25 & 52.6\%$\pm$2.7 \\
    7 & 9.32$\pm$ 0.55 & 50.3\%$\pm$6.3 \\
    6 & \textbf{8.98 $\pm$ 0.34}& \textbf{53.3\%$\pm$3.0} \\
    5 & 9.26$\pm$ 0.54 & 50.1\%$\pm$5.5\\
    \hline
    \end{tabular}
    \end{center}
    \end{table}

    \begin{table*}[t]
    \setlength\tabcolsep{3pt}
    \begin{center}
    \caption{The MAE and $\mathbf{R}^2$ results on synthetic dataset, and model performance improvement on this dataset is observed as stable as that on IXI dataset, confirming the stablity of the proposed GCLSS method.}
    \label{tab:synthetic-mae}
    \begin{tabular}{  c l | l | cccc | cccc  }
    \cline{1-11}		
    \multirow{9}{*}{\rotatebox{90}{Synthetic dataset}}&\multirow{2}{*}{Type}&\multirow{2}{*}{Method}& \multicolumn{4}{c|}{MAE $\downarrow$}&\multicolumn{4}{c}{$\mathbf{R}^2$$\uparrow$}  \\
    \cline{4-11}
    &&&1/5labels&1/4labels&1/3labels&1/2labels &1/5labels&1/4labels&1/3labels&1/2labels   \\ 
    \cline{2-11}	
    &\textit{Supervised}&Regression&0.098$\pm$0.095&0.056$\pm$0.016&0.041$\pm$0.015&0.032$\pm$0.009&66.9\%$\pm$39.4&83.8\%$\pm$7.7&88.5\%$\pm$8.0&90.9\%$\pm$5.1\\
    \cline{2-11}
    &&Mean-Teacher~\cite{tarvainen2017mean}&0.080$\pm$0.089&0.047$\pm$0.021&0.043$\pm$0.019&0.029$\pm$0.011&69.4\%$\pm$40.1&86.9\%$\pm$8.4&90.7\%$\pm$7.6&92.5\%$\pm$7.2\\
    \cline{3-11}
    &\textit{Semi-}&CPS~\cite{chen2021semi}&0.057$\pm$0.012&0.045$\pm$0.016&0.041$\pm$0.015&0.028$\pm$0.007&84.5\%$\pm$8.8&88.8\%$\pm$8.5&88.5\%$\pm$8.0&93.3\%$\pm$5.3\\
    \cline{3-11}
    &\textit{Supervised}&UCVME~\cite{dai2023semi}&0.040$\pm$0.008&0.033$\pm$0.008&0.027$\pm$0.007&0.028$\pm$0.021&92.2\%$\pm$3.6&94.2\%$\pm$2.8&95.0\%$\pm$3.0&95.6\%$\pm$4.3\\
    \cline{3-11}
    &&CLSS~\cite{dai2024semi}&\underline{0.033$\pm$0.008}&\underline{0.027$\pm$0.009}&\underline{0.020$\pm$0.007}&\underline{0.016$\pm$0.007}&\underline{96.4\%$\pm$1.7}&\underline{97.3\%$\pm$2.4}&\underline{98.4\%$\pm$1.3}&\underline{99.3\%$\pm$0.5}\\
    \cline{3-11}
    \cline{3-11}
    \cline{3-11}
    &&GCLSS&\textbf{0.024$\pm$0.009}&\textbf{0.021$\pm$0.004}&\textbf{0.014$\pm$0.003}&\textbf{0.010$\pm$0.003} &\textbf{97.8\%$\pm$1.4}&\textbf{98.4\%$\pm$0.6}&\textbf{99.4\%$\pm$0.2}&\textbf{99.7\%$\pm$0.1}\\ 
    \cline{1-11}
    \end{tabular}
    \end{center}
    \end{table*}

    However, this process assumes that the patients used for training have similar chronological and biological ages, requiring data exclusively from healthy individuals. By leveraging unlabeled data in the training process, we can enhance the robustness of the model. This is achieved by relaxing the strict enforcement of chronological age as the ground truth for these samples, thereby mitigating the reliance on a large pool of healthy patient data. Next we will concretely introduce the dataset and experimental settings.

    \textit{Dataset:} The IXI dataset comprises $588$ MRI brain volumes, each associated with an age ranging from $20$ to $86$ years. For experiments, we allocate 80 samples as the validation set and 88 samples as the test set. Our backbone model is the 3D ResNet-18~\cite{tran2018closer}. 

    \textit{Settings:} During training, we employ a learning rate of $1\times {10}^{-3}$, with a decay of $0.1$ every $10$ epochs, and train models for totally $30$ epochs. The batch size for labeled samples is set to $16$ considering the memory, and the batch size of unlabeled samples is set to $8$ with an ablation experiment. For the balancing hyper-parameters, we assign weights to the different components of loss functions as: ${\lambda}_{SC} = 0.1$, ${\lambda}_{UC}=0.05$, ${\lambda}_{UR}=0.01$, and $\lambda = 2$, and these values are carefully selected with additional ablation experiments. To account for training instability, we conduct separate experiments $8$ times, each with different random seeds. The reported results include the mean and standard deviation across these $8$ runs.

    \subsubsection{Comparisons with sota methods and ablation studies}
    For quantitative comparisons, we exhibit the results of GCLSS and other sota semi-supervised methods in Table.~\ref{tab:ixi}, including the conventional mean-teacher~\cite{tarvainen2017mean} (Mean-teacher) and cross psuedo-label supervision~\cite{chen2021semi} (CPS) semi-supervised learning methods for deep regression using a single output value as the regression prediction. Besides, we compare with the recent UCVME~\cite{dai2023semi} method, which enforces uncertainty consistency, and CLSS method~\cite{dai2024semi}. For performance reference, we additionally show results using a supervised naive regression method only with labeled data. Semi-supervised settings are employed where $1/5$, $1/4$, $1/3$, and $1/2$ of the data is accessible as labeled data and remaining samples are treated as unlabeled data. Obviously, we observe that CLSS and GCLSS both perform the best almost across all settings, and GCLSS further outperforms CLSS $4\%-7.6\%$ under the $\mathbf{R}^2$ metric. This confirms that the ground-truth label distance information from labeled samples regularizes the ordinal ranking obtained with mixed data, thus benefiting the regression results. GCLSS therefore can be used as an effective SSL regression method which reduces reliance on labeled data for medical imaging analysis applications. Next we introduce ablation experiments and demonstrate the importance of each component in GCLSS. Note that considering the complex relation between the batch size of unlabeled samples and the total number $B'$ of selected features in MFSM, we firstly set that MFSM totally selects the whole unlabeled batch when ablating the batch size of unlabeled samples and other hyper-parameters for simplicity, followed by ablating $B'$ of MFSM.
    
    \textit{Impact of unlabeled batch size:} To select the best suitable batch size for unlabeled samples, we test the sensitivity of our method to the size of the unlabeled batch. Particularly we show results using $1/5$ available labels with batch sizes $\{4, 8, 16, 32\}$ in Table.~\ref{table:batchsize-ixi}. We observe that the  performance improves when we gradually enlarge the batch size and the best performance is achieved when it is set to be $8$, where the ordinal ranking regularization is enforced with more information.
    Further enlarging the batch size provides more comprehensive data information, while it simultaneously shrinks the theoretically perturbation tolerances in Eq.~\ref{eq:deltaS} (and Eq.~\ref{eq:deltaz}) and thereby complicates the ordinal ranking process. These combined effects not only challenge model optimization but may ultimately undermine training stability, suggesting there exists an optimal and moderate batch size threshold for the best performance.
    Thus we use the batch size $8$ to perform GCLSS and enforce supervision on unlabeled samples. Note that we use the same number of labeled samples to compute ${\mathcal{L}}^{m}$, which are randomly chosen from the batch of labeled samples.

    \textit{Hyper-parameter sensitivity:}
    \label{sec:hyper}
    We next test the sensitivity of our method to each weight ${\lambda}_{SC}, {\lambda}_{UC}$, and ${\lambda}_{UR}$ in the loss function to select the best suitable hyper-parameters. We show results performed using $1/5$ available labels for reference in Table.~\ref{table:hypers-ixi}. Each parameter is adjusted individually in a small range whilst keeping the remaining ones at the optimum value. Taking ${\lambda}_{SC}$ as an example, we test model performances with candidates $[1.0, 0.1, 0.01]$ considering the scale of ${\mathcal{L}}_{SC}$, and we observe that our model performs the best with $0.1$. Similarly, we obtain the best suitable value of ${\lambda}_{UC}$ and ${\lambda}_{UR}$ at $0.05$ and $0.01$, respectively. Thus we fix them in later experiments.

    \textit{Impact of $B'$:} For the introduced memory-based feature selection module, where the total number of selected features plays a crucial role in learning, we conduct sensitivity tests of the GCLSS algorithm to examine the impact of different values for $B'$. Specifically, we fix the best suitable unlabeled batch size to $8$ and other hyper-parameters and evaluate the performance for various settings of $B'$, namely $[8, 7, 6, 5]$, as shown in Table.~\ref{table:B'}. It is important to note that the values gradually decrease from the chosen unlabeled batch size. Our observations reveal that the model achieves the optimal performance when $B'$ is set to 6. Larger choices result in increased variations, while smaller choices provide limited ranking information. Hence, we set $B'$ to 6 for subsequent experiments conducted on the IXI dataset.

    \begin{table}[t!]
    \fontsize{9pt}{10.8pt}\selectfont
    \begin{center}
    \caption{We compare the model parameters and computation time between our proposed method and involved methods on the synthetic dataset.}
    \label{table:paramsandtime}
    \begin{tabular}{l |l |c |c}
    \hline
    Type &Method & \# Params &  Infer Time \\ 
    \hline
    \textit{Supervised}  & Regression  &34,401 & 0.0013 \\
    \hline
                         &Mean-teacher &68,802    & 0.0012 \\
    \textit{Semi-}       &CPS          &68,802    & 0.0018 \\
    \textit{Supervised} &UCVME        &69,004    & 0.0043 \\
                         &CLSS         &34,401    & 0.0013 \\
                         &GCLSS        &34,401    & 0.0013 \\
    \hline
    \end{tabular}
    \end{center}
    \end{table}

    \begin{table*}
    \begin{center}
    \caption{The MAE and $\mathbf{R}^2$ results on AgeDB-DIR dataset.}
    \label{tab:agedb}
    \begin{tabular}{  c l | l | cccc  }
    \cline{1-7}
    \multirow{8}{*}{\rotatebox{90}{AgeDB-DIR}}&\multirow{2}{*}{Type}&\multirow{2}{*}{Method}&\multicolumn{4}{c}{MAE $\downarrow$}    \\ 
    \cline{4-7}
    &&&1/30labels&1/25labels&1/20labels&1/15labels  \\ 
    \cline{2-7}
	&\textit{Supervised}&Regression&10.14$\pm$0.25&9.99$\pm$0.11&9.10$\pm$0.15&8.58$\pm$0.10 \\
    \cline{2-7}
    &&Mean-Teacher~\cite{tarvainen2017mean}&10.05$\pm$0.29&9.99$\pm$0.13&9.05$\pm$0.12&8.62$\pm$0.09 \\
    \cline{3-7}
    &\textit{Semi-}&CPS~\cite{chen2021semi}&9.99$\pm$0.12&9.83$\pm$0.10&8.99$\pm$0.14&8.47$\pm$0.08 \\
    \cline{3-7}
    &\textit{Supervised}&RankUp~\cite{huang2024rankup}&\textbf{9.84$\pm$0.10}&9.76$\pm$0.11&9.00$\pm$0.08&8.52$\pm$0.12 \\
    \cline{3-7}
    &&CLSS~\cite{dai2024semi}&9.95$\pm$0.18&\underline{9.59$\pm$0.12}&\underline{8.88$\pm$0.09}&\underline{8.45$\pm$0.11} \\
    \cline{3-7}
    \cline{3-7}
    &&GCLSS&\underline{9.86$\pm$0.17}&\textbf{9.50$\pm$0.14}&\textbf{8.77$\pm$0.09}&\textbf{8.42$\pm$0.11} \\
    \cline{1-7}
    \end{tabular}
    \end{center}
    \end{table*}

    \begin{table*}[h]
    \begin{center}
    \caption{The MAE and $\mathbf{R}^2$ results on UTK dataset.}
    \label{tab:utk}
    \begin{tabular}{  c l | l | ccc | ccc  }
    \cline{1-9}	
    \multirow{9}{*}{\rotatebox{90}{UTK dataset}}&\multirow{2}{*}{Type}&\multirow{2}{*}{Method}& \multicolumn{3}{c|}{MAE $\downarrow$}&\multicolumn{3}{c}{$\mathbf{R}^2$$\uparrow$}  \\
    \cline{4-9}
    &&&1/30labels&1/20labels&1/10labels &1/30labels&1/20labels&1/10labels  \\ 
    \cline{2-9}
    &\textit{Supervised}&Regression&7.92$\pm$0.12&6.21$\pm$0.12&5.69$\pm$0.09&27.1\%$\pm$1.5&43.8\%$\pm$7.5&51.0\%$\pm$3.1\\
    \cline{2-9}
    &&Mean-Teacher~\cite{tarvainen2017mean}&7.04$\pm$0.08 &6.15$\pm$0.08&5.54$\pm$0.07&32.6\%$\pm$1.3 &45.7\%$\pm$1.1&54.3\%$\pm$0.4 \\
    \cline{3-9}
    &&CPS~\cite{chen2021semi}&6.99$\pm$0.10&5.92$\pm$0.07&5.40$\pm$0.03&34.8\%$\pm$1.5&47.9\%$\pm$1.1&56.3\%$\pm$0.5 \\
    \cline{3-9}
    &\textit{Semi-}&UCVME~\cite{dai2023semi}&6.43$\pm$0.08&5.84$\pm$0.06&\underline{5.26$\pm$0.02} &41.4\%$\pm$0.7&49.4\%$\pm$0.7&57.9\%$\pm$0.3\\
    \cline{3-9}

    &\textit{Supervised}&RankUp~\cite{huang2024rankup}&6.32$\pm$0.02&5.92$\pm$0.02&5.59$\pm$0.01&35.9\%$\pm$0.5&47.8\%$\pm$0.3&50.6\%$\pm$0.2\\
    \cline{3-9}

    &&CLSS~\cite{dai2024semi} &\underline{6.19$\pm$0.04} &\underline{5.79$\pm$0.03} &\underline{5.26$\pm$0.02} &\underline{44.3\%$\pm$0.6} &\underline{51.1\%$\pm$0.4} &\underline{58.0\%$\pm$0.4} \\
    \cline{3-9}
    &&GCLSS&\textbf{6.18$\pm$0.03}&\textbf{5.72$\pm$0.02} & \textbf{5.22$\pm$ 0.02} &\textbf{45.4\%$\pm$0.5}&\textbf{52.0\%$\pm$0.5} &\textbf{58.3\%$\pm$0.2} \\
    \cline{1-9}
	
    \end{tabular}
    \end{center}
    \end{table*}

    \textit{Impact of generalized seriation ranking and feature selection:} Our proposed GCLSS indeed differs from CLSS in two aspects: 1). the current ordinal ranking contains regularization from labeled data, i.e., the guidance of ground-truth information; 2). the introduced memory-based feature selection module helps reduce the feature variation. In this case, to verify the effectiveness of the two introduced techniques, we conduct ablation experiments still using $1/5$ available labels and show results in Table.~\ref{tab:ablation-module}. It is observed that the model performance drops a lot when we get rid of either technique, while both performances are still better than CLSS. This confirms that the two techniques both contribute to our final performances and using each only would damage the effect.

    \subsection{Validation on synthetic dataset for non-linear operator learning}
    \textit{Dataset:} We next use the non-linear synthetic dataset~\cite{lu2021learning} and train a neural network to estimate the operator function. The target is the stochastic partial differential equation:
    \begin{equation*}
        - \text{div} ({e}^{b(z;\omega)}\triangledown u(z;\omega)) = f(z), \quad z\in(0, 1) \text{ and } \omega\in \Omega
    \end{equation*}
    with Dirichlet boundary conditions $u(0) = u(1) = 0$, and $f (z) = 10$. ${e}^{b(z;w)}$ is a diffusion coefficient with $b(z;w)$ following a random Gaussian process. The input data $ \{ {x}_{i} \}_{i=1}^{N} $ are outputs generated by $b(z; w)$ and the target label $ \{ {y}_{i} \}_{i=1}^{N} $ is the solution of $u(z; w)$~\cite{lu2021learning}.

    \textit{Settings:} For the network structure and training scheme, following~\cite{zhang2022improving}, we employ a two-layer fully connected neural network with $100$ hidden units as the backbone. We perform $10$ separate training runs on $1,000$ samples each and test our model on a set of $100,000$ samples. Mean and standard deviation of the $10$ runs are reported in Table.~\ref{tab:synthetic-mae}. The learning rate is set to be $1\times {10}^{-3}$. We use the entire dataset as an input batch and train for $100,000$ epochs. While considering the quadratic scaling of the feature similarity matrix, a smaller batch size of $8$ samples are used to calculate the loss function to reduce computation time. The balancing hyper-parameters ${\lambda}_{SC}$, ${\lambda}_{UC}$, and ${\lambda}_{UR}$ are set to be $1{e}^{-3}$, $1{e}^{-3}$ and $1{e}^{-4}$, which are selected following the same ablation precedures as in Section.~\ref{sec:hyper}.

    \subsubsection{Comparisons with sota methods} 
    We compare with the previously introduced sota semi-supervised deep regression methods, Mean-Teacher, CPS, UCVME, and CLSS, to demonstrate the effectiveness of GCLSS. Performances of a naive supervised regression method using only labeled data are also shown for reference. The specific results are shown in Table.~\ref{tab:synthetic-mae} for different settings, where $1/5$, $1/4$, $1/3$, and $1/2$ of available labels are used, and the remaining samples are treated as unlabeled data. Obviously, semi-supervised methods generally outperform naive supervised regression. Further, our proposed method, GCLSS, convincingly outperforms alternative methods and consistently improves $\mathbf{R}^2$ by 1-1.4\% over the next best alternative across all settings except for $1/2$ setting, where the CLSS is accurate enough. 
    
    \subsubsection{Computation analysis} To explore whether our proposed method introduces additionally significant computational complexity, we next provide the total number of parameters and reference time comparisons in seconds for performing one iteration of inference in Table.~\ref{table:paramsandtime}. Practical times may differ depending on various hardware and environment. Obviously our inference time is comparable or more efficient than sota semi-supervised methods because we only require predictions from one model, instead of taking the average from two co-trained models. For the total number of model parameters required for each method, we note that GCLSS only uses one model, whilst some alternative methods, e.g., CPS and UCVME, rely on two co-trained models which requires more memory. Besides, we declare that both CLSS and GCLSS do not introduce significant computational complexity in training since additional calculations involving eigenvalue decomposition can be performed efficiently with existing computational tools and algorithms.

    \subsection{Validation on age-estimation from photography}
    To further substantiate our approach, we have conducted additional validation on two natural image datasets. Specifically, we utilize the AgeDB-DIR dataset~\cite{yang2021delving} and UTKFace dataset~\cite{zhang2017age} for age estimation from natural photographs. These datasets serve as widely recognized benchmarks for deep regression tasks in age estimation. While personal photographs can be readily obtained using various techniques, obtaining accurate age labels is often hindered by privacy concerns. This challenge can be effectively addressed through the application of semi-supervised deep regression methods, which reduces the dependence on labeled data and enhances the robustness of age estimation models. Note that we additionally compare with the sota RankUp~\cite{huang2024rankup} on these two datasets as the method is originally experimented and performs comparable for the age estimation task.
    
    \subsubsection{AgeDB-DIR Dataset}
    
    The AgeDB-DIR dataset consists of $16,488$ personal photographs with ages ranging between $1$ and $101$, and has fewer tail samples to reflect real-world label imbalance phenomenon. We use the same data splits as provided by the dataset for training, validation, and testing~\cite{yang2021delving}.

    \textit{Settings:} For our deep regression backbone model, we adopt the pretrained network ResNet50~\cite{he2016deep} on the ImageNet dataset~\cite{deng2009imagenet}. During training, we set the learning rate to $5\times {10}^{-4}$, with a decay of $0.1$ every $10$ epochs. The models are trained for a total of $30$ epochs. We utilize a batch size of $32$ for labeled samples and a batch size of $8$ for unlabeled samples to ensure efficient training. The hyper-parameters ${\lambda}_{SC}$, ${\lambda}_{UC}$, ${\lambda}_{UR}$, and $\lambda$ are set to be $1$, $0.05$, $0.01$, and 2, which are determined through extensive ablation experiments similar to the process described for the IXI dataset. To address training instability, we conduct separate experiments $8$ times, each with different random seeds. This approach enables us to account for potential variations in results. The reported outcomes include the mean and standard deviation across these $8$ runs, providing a comprehensive assessment of the model performance.

    \textit{Result analysis:} To verify the performance of our proposed GCLSS method, we compare the results with sota semi-supervised methods, as presented in Table.~\ref{tab:agedb}. We evaluate with different settings where varying fractions of the dataset, specifically $1/30$, $1/25$, $1/20$, and $1/15$, are treated as labeled data, while the remaining samples are treated as unlabeled data. Our findings consistently demonstrate that both CLSS and GCLSS outperform the alternative methods across all settings, except that RankUp performs better when $1/30$ labels are available. While we observe that the performances of RankUp is not as stable as CLSS and GCLSS, especially when the available labels increase. We attribute this phenomenon to RankUp's effective utilization of unlabeled data while failing to leverage the comparative information embedded in inter-sample relationships (i.e., labeled-labeled and labeled-unlabeled sample comparisons). Furthermore, GCLSS exhibits further improvements over CLSS, indicating its enhanced efficacy. These results substantiate that our GCLSS methods can be effectively applied to natural image datasets. Importantly, the incorporation of regularization from labeled data and the utilization of MFSM have proven to be highly effective, even with varying scene diversity and image intensity.

    \subsubsection{UTK Dataset}
    The UTKFace dataset is consisted of totally $13,144$ photographs available for training and $3,287$ photographs for testing. Here we use a subset of the training dataset for validation and follow the split in previous works~\cite{cao2020rank}. Faces have been pre-cropped and age labels range from $21$ to $60$.

    \textit{Settings:} We also use the ResNet50~\cite{he2016deep} as our encoder and add additional dropout layers after each of the four main residual blocks. The model is trained for 30 epochs using the learning rate $1\times {10}^{-4}$ with weight decay ${10}^{-3}$ and the Adam optimizer. We use a batch size of $32$ for labeled data and $8$ for unlabeled data. To account for unstable training, all experiments are run separately $8$ times using different random seeds. Mean and standard deviation of the $8$ runs are reported.
    
    \textit{Result analysis:} 
    For the validation of the performance of our proposed GCLSS method on the dataset, we compare with sota semi-supervised methods, as in Table.~\ref{tab:utk}. These comparisons are conducted using different configurations, where varying proportions of the dataset, specifically $1/30$, $1/20$, and $1/10$, are utilized as labeled data, while the remaining samples are considered as unlabeled data. Consistently across all configurations, our findings highlight the superiority of GCLSS over the alternative methods. Notably, GCLSS showcases further enhancements over ordinal ranking-based CLSS and RankUp, underscoring its superior effectiveness with the regularization additionally provided by mixed matrix ${\mathcal{S}}^{m}$. These results provide compelling evidence that our CLSS and GCLSS methods are well-suited for effective application to natural image datasets.
    
    \begin{table}[t!]
    \fontsize{9pt}{10.8pt}\selectfont
    \begin{center}
    \caption{The MAE and $\mathbf{R}^2$ results on BVCC dataset. To ensure consistency with our previous hardware environment, we have re-implemented the preprocessing pipeline and experiments reported in the RankUp method within our own environment, adopting identical hyper-parameter settings. Our results generally outperform those reported in the RankUp's table, and our GCLSS still achieves the best performance.}
    \label{tab:bvcc}
    \begin{tabular}{l |l |c |c}
    \hline
    Type &Method & MAE $\downarrow$  & $\mathbf{R}^2$$\uparrow$\\ 
    \hline
    \textit{Supervised}  & Regression  &0.527$\pm$0.001  & 50.4\%$\pm$0.5 \\
    \hline
    &Mean-teacher~\cite{tarvainen2017mean} 
    &0.515$\pm$0.007  & 52.4\%$\pm$1.2 \\
    \cline{2-4}
    \textit{Semi-}&UCVME~\cite{dai2023semi}  &0.489$\pm$0.004  & 57.2\%$\pm$0.7  \\
    \cline{2-4}
    \textit{Supervised} &RankUp~\cite{huang2024rankup}        &\underline{0.460$\pm$0.005}  & \underline{60.1\%$\pm$1.0} \\
    \cline{2-4}
    &CLSS~\cite{dai2024semi} &0.465$\pm$0.005  & 59.4\%$\pm$0.8 \\
    \cline{2-4}
    &GCLSS            & \textbf{0.453$\pm$0.005} &\textbf{61.2\%$\pm$0.7}  \\
    \hline
    \end{tabular}
    \end{center}
    \end{table}
    
    \subsection{Generalizability validation on audio quality assessment}
    Although we have validated the semi-supervised regression performances of GCLSS on both synthetic and real-world datasets across diverse scenarios, the input modality has so far been limited to images. To further demonstrate the generalizability of our method to other modalities, we also evaluate it on the BVCC~\cite{cooper2021voices} dataset for audio quality prediction. Composed of high-quality audio samples in multiple languages, this dataset is widely regarded as a benchmark for audio quality assessment. We therefore adopt it as a complementary modality to verify the effectiveness of GCLSS in semi-supervised regression.

    \textit{Dataset:} 
    The BVCC dataset is divided into $4,974$ audio samples for training, $1,066$ for evaluation, and another $1,066$ for testing. The corresponding labels are mean opinion scores (from $1$ to $5$) derived from multiple listener ratings. In our experiments, we follow the dataset split protocol of RankUp~\cite{huang2024rankup}, which designates the original $4,974$ samples for  training and the $1,066$ evaluation samples for validation.
    
    \textit{Settings:} 
    Since the audio samples differ from images, we adopt the pretrained Whisper-base~\cite{radford2023robust} model to extract features. The model is trained for $102400$ iterations using the learning rate $2\times{10}^{-6}$ with weight decay $2\times {10}^{-5}$ and the AdamW optimizer. We use a batch size of $8$ for labeled data and $8$ for unlabeled data. The involved hyper-parameters ${\lambda}_{SC}$, ${\lambda}_{UC}$, ${\lambda}_{UR}$, and $\lambda$ are set to be $1$, $0.05$, $0.01$, and $2$. To account for training instability, all experiments are run separately $6$ times using different random seeds, and mean and standard deviation of the $6$ runs are reported.

    \textit{Result analysis:} 
    To validate the assessment performances of GCLSS, we also compare it with other sota methods and present results in Table.~\ref{tab:bvcc}. The experiment is conducted using $250$ labeled audio samples (approximately $1/20$ of the data), with the remaining samples treated as unlabeled. Under this setting, we again observe consistent superior performance over other methods. Notably, all three ordinal-ranking-based approaches achieve satisfactory results, and our GCLSS outperforms both RankUp and CLSS, benefiting from feature-level contrastive regularization that incorporates both labeled and unlabeled samples. The results further proves that our GCLSS method generalizes effectively to sequential data.

\section{Discussion and conclusion}


In this paper, we introduce GCLSS, a novel semi-supervised regression method that leverages unsupervised contrastive learning to utilize unlabeled data for establishing relative sample distance rankings. By constructing a feature similarity matrix using both labeled and unlabeled samples and employing a spectral seriation algorithm, we obtain robust rankings for unlabeled samples, which are treated as pseudo-supervision. These rankings, obtained while regularized by the ground-truth label information from labeled samples, ensure reliability and benefit the regression task. Theoretical bounds for error values in the similarity matrix and features demonstrate the robustness of GCLSS. Through comprehensive empirical evaluations on diverse datasets, including medical image analysis, synthetic operator learning, and natural image datasets, we show that GCLSS outperforms alternative state-of-the-art semi-supervised deep regression methods. Our findings highlight GCLSS as a valuable technique for improving the performance of semi-supervised deep regression methods, enabling the effective utilization of unlabeled data and contrastive learning principles.

\bibliographystyle{IEEEtran} 
\bibliography{general_ref}

\end{document}